\documentclass[runningheads]{llncs}
\usepackage{graphicx}
\usepackage[marginal]{footmisc}

\usepackage{bbding}
\usepackage{amsmath}
\usepackage{booktabs}
\usepackage{threeparttable} 
\usepackage{multirow}
\usepackage{parskip}
\usepackage{subfigure}

\DeclareUnicodeCharacter{1EA7}{}
\begin{document}
\title{FLAME-based Multi-View 3D Face Reconstruction}
\author{Wenzhuo Zheng\thanks{Wenzhuo Zheng and Junhao Zhao contribute equally to this work.} \and
Junhao Zhao\textsuperscript{$*$} \and
Xiaohong Liu\inst{(}\Envelope\inst{)} \and
Yongyang Pan \and
Zhenghao Gan \and
Haozhe Han \and
Ning Liu\inst{(}\Envelope\inst{)}
}
\authorrunning{W. Zheng et al.}
% \institute{Shanghai Jiao Tong University, Shanghai, China \and
\institute{Shanghai Jiao Tong University, Shanghai, China\\
\email{\{darkcorvus,200217zjh,xiaohongliu,panyongyang,\\ ganzhenghao,h2411522561,ningliu\}@sjtu.edu.cn}}
\maketitle              % typeset the header of the contribution
\footnote{Supported in part by Shanghai Pujiang Program under Grant 22PJ1406800 and Shanghai Jiao Tong University under U1908210.\\}
\begin{abstract}
At present, face 3D reconstruction has broad application prospects in various fields, but the research on it is still in the development stage. In this paper, we hope to achieve better face 3D reconstruction quality by combining a multi-view training framework with face parametric model FLAME, and propose a multi-view training and testing model \textbf{MFNet} (Multi-view FLAME Network). We build a self-supervised training framework and implement constraints such as multi-view optical flow loss function and face landmark loss, and finally obtain a complete MFNet. We propose innovative implementations of multi-view optical flow loss and the covisible mask. We test our model on AFLW and facescape datasets and also take pictures of our faces to reconstruct 3D faces while simulating actual scenarios as much as possible, which achieves good results. Our work mainly addresses the problem of combining parametric models of faces with multi-view face 3D reconstruction and explores the implementation of a FLAME-based multi-view training and testing framework for contributing to the field of face 3D reconstruction.

\end{abstract}
\section{Introduction}
% Please note that the first paragraph of a section or subsection is
% not indented. The first paragraph that follows a table, figure,
% equation etc. does not need an indent, either.

% Subsequent paragraphs, however, are indented.
\label{sec:1}
Face 3D reconstruction~\cite{roberts1963machine} mainly focuses on the reconstruction of human facial regions,
and broadly speaking, also includes hair, ear, neck, and other regions. The human face is a special 3D object that has not only more complex shape and texture features, but also strong prior constraints. This poses a great challenge to face 3D reconstruction on one hand, and on the other hand, it also provides feasible technical approaches to reconstruct the face 3D structure from 2D information, and the face parametric model is one of them. The face parametric model is a statistical model based on a large number of faces, and its core idea is that faces can be matched one-to-one in the 3D feature space and can be obtained by weighted linear summation of orthogonal bases for a large number of other faces. The most widely used model is 3DMM~\cite{blanz1999morphable,blanz2003face}, but it has two core problems: (1) 3DMM is in a low-dimensional space and thus the face detail characterization is weak; (2) 3DMM only reconstruct the front face region without neck or hindbrain. Therefore, we choose FLAME~\cite{li2017learning}, which has a better characterization of details and more complete reconstruction. FLAME has three parameters: shape, pose, and expression, which can more accurately classify faces into more dimensions, and the face reconstructed by FLAME includes the whole head. However, there is not much research work on FLAME so far, and there is a gap in the field of multi-view training using FLAME. Our work fills this gap and makes an exploratory contribution to FLAME-based multi-view training.

In the past decade, deep learning technologies have become a dominant trend in face 3D reconstruction. Some works~\cite{feng2021learning,sanyal2019learning} use neural networks to regress end-to-end to compute the inputs needed for face parameterization models, but are limited to single-view, while our proposed MFNet can utilize features from multiple views and fuse them to obtain more complete face information. In this paper, we use FLAME as a powerful tool to reconstruct fine-grained 3D face models with low cost and only 2D RGB images.

Our main contributions are listed as follows:
\begin{itemize}
  \item We innovatively combine multi-view training with FLAME, propose a multi-view self-supervised framework, and implement a complete multi-view training and testing process. Our proposed model MFNet achieves good results on both test datasets and actual captured images.
  
  \item We propose a multi-view optical flow loss for our multi-view training framework and propose a novel implementation of the technical details such as covisible mask.
\end{itemize}

\section{Related Work}
\label{sec:2}

\subsection{Parametric model}
In 1999, Blanz and Vetter et al.~\cite{blanz1999morphable,blanz2003face} proposed the 3D Morphable Model (3DMM) for the human face, which is the most widely used 3D face reconstruction model. Subsequent studies related to 3DMM have been published in the next decade, either by adding coefficients to the original model, such as Pascal Paysan et al.~\cite{gerig2018morphable} updated the expression coefficients of the 3DMM model for BFM (Basel Face Model) model in 2017, or build larger datasets, such as James Booth et al.~\cite{booth20163d} built a dataset of 9663 faces, or propose better ways to optimize the solution coefficients, such as adding deep learning ideas to the coefficient solution in recent years to achieve better results ~\cite{booth20173d,tran2019learning}, or make nonlinear adjustments to the model, such as the nonlinear 3DMM model proposed by Luan Tran et al. ~\cite{tran2018nonlinear}, but none of them have departed from the original framework of 3DMM. This also leads to the fact that these changes do not solve the two core problems of 3DMM mentioned above. Therefore, we choose FLAME\cite{li2017learning} as our face parametric model.

FLAME was proposed by Li Tianye et al., referring to the expression of the body model SMPL\cite{loper2015smpl}, combining linear blend skinning (LBS) and the corresponding corrected blendshape. Not many researches have been done on FLAME\cite{feng2021learning,sanyal2019learning}, and they are all limited to single-view. We want to utilize the features and data from multiple perspectives, so we propose a self-supervised multi-view training framework and achieve better reconstruction results.

\subsection{Multi-view reconstruction}

There are many works based on face parametric models, but very few of them\cite{shang2020self,wu2019mvf} are trained using multi-view data, and the only ones are based on 3DMM. MVFNet\cite{wu2019mvf} is the first work that proposed the idea of multi-view parametric model training, but it is based on 3DMM and the implementation is very rough, which leads to poor results. MGCNet\cite{shang2020self} makes some improvements on its basis, proposing novel multi-view loss functions, using multi-view training, but only using a single image for testing. It does improve the quality of the face reconstruction, but the reconstructed faces were still rough and incomplete. The field of FLAME-based multi-view training remains a gap. To the best of our knowledge, MFNet is the first work on 3D face reconstruction using multi-view training and testing framework based on the face parametric model FLAME.

\section{Method}
\label{sec:3}

\subsection{Overall architecture}

\begin{figure*}[!htp]
  \centering
  \includegraphics[width=0.95\textwidth]{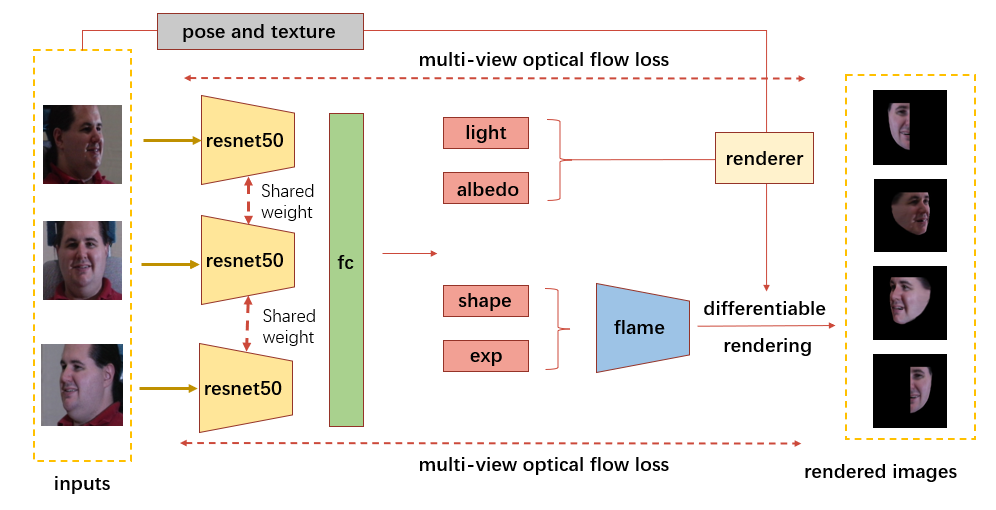}
  \caption{Architecture of MFNet.}
 \label{fig: overview}
\end{figure*}

The overall architecture for our proposed is show in Figure~\ref{fig: overview}. Resnet is a highly mature technology that has performed well in numerous image recognition and classification. So we extract features from each input image by a shared weight Resnet50, and then concatenate the features together and put them into a fully connected layer to regress a set of flame parameters for the person. Also, we separate a pose and texture feature from Resnet50 for each perspective for subsequent reconstruction work and calculate loss.

\subsection{FLAME}
\label{sec:FLAME}

After extracting features from the multi view images in the input batch through Resnet50 and converting them into fully connected layers, we can obtain the desired FLAME model input vectors $\vec{\beta}$, pose $\vec{\theta}$, expression $\vec{\psi}$. Next, the FLAME model acts as a decoder to convert these hidden layer vectors into three-dimensional facial information.These three-dimensional information mainly consists of two parts, the first is the information of each vertex, such as coordinate $T_P$, Normal vector $N_{uv}$ and faces $F$, and the second is landmark coordinates of the face. The equation of the FLAME model is as follows:

\begin{equation}
  M(\vec{\beta}, \vec{\theta}, \vec{\psi}) = W(T_P(\vec{\beta}, \vec{\theta}, \vec{\psi}), \mathbf{J}(\vec{\beta}), \vec{\theta}, \mathcal{W})
  \label{eq:FLAME}
\end{equation}
\subsection{Feature extraction}
\label{sec:Fea_extra}

We use part of DECA\cite{feng2021learning} as the pretrained model of Resnet50 for better feature extractoin and finetune it. In order to obtain better feature information, we use a fully connected layer to fuse the features extracted by Resnet50 from three perspectives together for consideration, thereby obtaining a more accurate model.

\subsection{Differentiable renderer}
\label{sec:diffrential}
After getting the 3D information of the face through FLAME model, we need to use 3D rendering to get the 2D image.Our shadow facial image $B(alpha, l, N_ {UV})$ is calculated based on the following equation:

\begin{equation}
  B(\alpha, l, N_{uv})_{i,j} = A(\alpha)_{i,j} \odot \sum\limits_{k=1}^9 l_k H_k(N_{i,j})
  \label{eq:buv}
\end{equation}

\noindent In the equation~\ref{eq:buv}, $A(\alpha)$ represents UV albedo map, $N_ {UV} $is the normal vector of the face surface output by FLAME. $B_{i,j} \in R^3$, $A_{i,j} \in R^3$, $N_{i,j} \in R^3$ represents the various attributes of pixel $(i, j)$ in the UV coordinate system. $\odot$ represents Hadamard product.

In addition, we also need to extract texture from the original input image and obtain vertex coordinates $T_P$ and faces $F$ to calculate the correspondence between points in the 3D mesh and the 2D texture map $U_ {V}$. Then, the texture map $I_{uv}^ \prime$ is obtained from the original input image by using this correspondence $U_{V}$, and the missing part in the middle is supplemented by bilinear interpolation. We extract the texture of multi views and perform simple fusion to obtain $I_{uv}^ \prime $, which contains information from multi views. Finally, we use facial mask $M_{face}$ to get UV texture map $I_{uv}$:

\begin{equation}
  I_{uv} = M_{face} \odot I_{uv}^ \prime 
  \label{eq:uv}
\end{equation}

\noindent Given the geometric parameters $(\vec{\beta}, \vec{\theta}, \vec{\psi})$, albedo $\alpha$, lighting condition $l$, and camera parameter $c$ of the mesh, we can render different two-dimensional face images $I_r$ from various perspectives:
\begin{equation}
  I_r = \mathcal{R}(M, B, c,I_{uv})
  \label{eq:render}
\end{equation}

\subsection{Loss function}
\label{sec:loss}
\subsubsection{Multiview optical loss}
\begin{figure}[!htpb]
  \centering
  \includegraphics[width=0.15\textwidth]{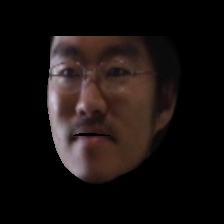}
  \hspace{1cm}
  \includegraphics[width=0.15\textwidth]{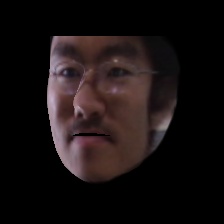}
  \hspace{1cm}
  \includegraphics[width=0.15\textwidth]{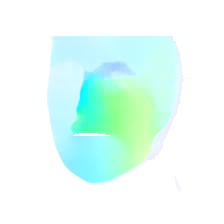}
  \caption{Optical flow estimation. From left to right are original image, rendered image and the optical flow. We use RAFT\cite{teed2020raft} to extract optical flow.}
  \label{fig:op}
\end{figure}

The optical flow loss\cite{guang2023acc}  calculates the optical flow between the rendered facial image and the original image. The design of the optical flow loss is based on an intuitive fact. That is, the coordinates of a point on a correct 3D model projected onto a 2D plane should be the same as the original image. We hope that these two points can coincide, so the distance should be as close to zero as possible. And that’s exactly what the optical flow loss does(Figure~\ref{fig:op}). 

However, due to the occlusion of the face, the reconstruction of the invisible part of the image view becomes very blurry. So we proposed an implementation method for a covisible mask. It can mask the blurry parts, so that these parts do not participate in the calculation of the optical flow loss. For the input face image, we first generate a projected two-dimensional face mask $MF$ according to the position of the three-dimensional face model. Then we use face landmarks to roughly extract the parts that can be seen from two viewpoints and get $MB$. The bounding box $MB$ composed of keypoints and the face mask $MF$ can be combined to obtain a better covisible mask $MC$:

\begin{equation}
  MC_{a,b} = MB_{a,b} \odot MF_b
\end{equation}

\begin{figure*}[!htpb]
  \centering
  \includegraphics[height=1.2cm, scale=0.5]{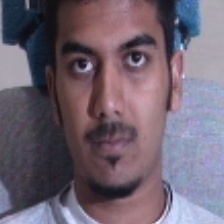}
  \hspace{1cm}
  \includegraphics[height=1.2cm, scale=0.5]{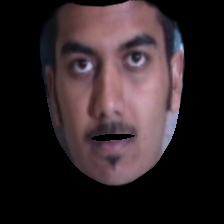}
  \hspace{1cm}
  \includegraphics[height=1.2cm, scale=0.5]{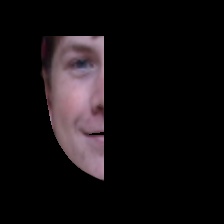}
  
  \includegraphics[height=1.2cm, scale=0.5]{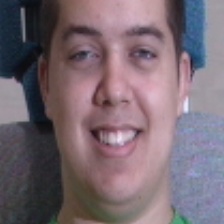}
  \hspace{1cm}
  \includegraphics[height=1.2cm, scale=0.5]{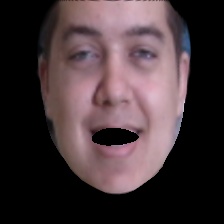}
  \hspace{1cm}
  \includegraphics[height=1.2cm, scale=0.5]{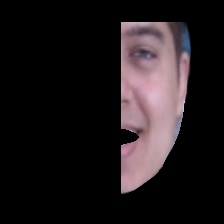}
  \caption{Covisible mask. From left to right are original images, rendered images and covisible masked images.}
  \label{fig:cov}
\end{figure*}

\noindent Here we show the usage of the covisible mask(Fig.~\ref{fig:cov}). In order to reduce the estimation error of the optical flow for the uninterested region, we also mask the complex regions such as the mouth, so that the covisible mask basically achieves our expected goal.

Given the image $I_b$ and the rendered image $I_{a \rightarrow b}$, the optical flow estimator $\mathbf{F}$, the covisible mask $MC_{a,b}$, we can calculate the multi-view optical flow loss function $L_{multiop}$:

\begin{equation}
  L_{multiop}(I_b, I_{a \rightarrow b}) = \lvert \mathbf{F}(MC_{a,b} \odot I_b, MC_{a,b} \odot I_{a \rightarrow b}) \rvert
\end{equation}
\subsubsection{Single View Keypoint Loss}

We project the 3D face keypoints to the 2D image and re-projecte them back to compared them. We hope that this can provide stronger face constraints for the model and prevent it from ignoring the constraints of the face itself:

\begin{equation}
  L_{singlelmk}(k_a, k_{a \rightarrow a}) = \sum\limits_{i \in MF_a} \Vert k_a(i) - k_{a \rightarrow a} (i) \Vert_1
  \label{eq:singlelmk}
\end{equation}

\subsubsection{Eye and lip keypoint loss}
Since the eye and lip area of the face is relatively complex, we implemented an eye keypoint loss and a lip keypoint loss to achieve better face reconstruction results. We compute the relative offset between the keypoints $k_a(i)$ and $k_a(j)$ of the upper and lower eyelids and lips on a certain view $a$, and measure the difference between their offset and the offset between the corresponding re-projected keypoints $k_{a \rightarrow a}(i)$ and $k_{a \rightarrow a}(j)$ of the 3D model:

\begin{equation}
    \begin{aligned}
          L_{eye}(k_a, k_{a \rightarrow a}) & = \\  \sum\limits_{(i,j) \in E_a} & \Vert k_a(i) - k_a(j) - ( k_{a \rightarrow a}(i) - k_{a \rightarrow a}(j) ) \Vert_1
          \label{eq:eye}
    \end{aligned}
\end{equation}

\begin{equation}
\begin{aligned}
        L_{lip}(k_a, k_{a \rightarrow a}) = & \\
        \sum\limits_{(i,j) \in P_a} \Vert k_a(i) & - k_a(j) - ( k_{a \rightarrow a}(i) - k_{a \rightarrow a}(j) ) \Vert_1
\label{eq:lip}
\end{aligned}
\end{equation}

\subsubsection{Regularized loss}
We need to regularize some vectors to prevent overfitting, including shape vector $\vec{\beta}$ regularization, expression vector $\vec{\psi}$ regularization and albedo $\alpha$ regularization:

\begin{equation}
  L_{reg} = \Vert \vec{\beta} \Vert_2 + \Vert \vec{\psi} \Vert_2 + \Vert \alpha \Vert_2 
  \label{eq:reg}
\end{equation}

\subsubsection{Total loss}
The total loss function is shown below:

\begin{equation}
\begin{aligned}
    L_{total} = 
    \lambda_1 L_{multiop} + \lambda_2 L_{singlelmk} + \lambda_3 L_{eye} + \lambda_4 L_{lip} +  \lambda_5 L_{reg}
  \label{eq:newtotal}
\end{aligned}
\end{equation}

\section{Experiments}
\label{sec:4}
In this section, we first introduce our implementation details for conducting the experiments, including the datasets and evaluation metrics(Sec. \ref{sec:detail}). Then we make qualitative and quantitative comparisons to other 3D face reconstruction methods(Sec. \ref{sec:qualitative} and Sec. \ref{sec:quantitative}). Finally, we demonstrate the effectiveness of the proposed method with extensive ablation studies in Sec. \ref{sec:ablation}.

\subsection{Implementation Details}
\label{sec:detail}

\subsubsection{Training Datasets} Our training is performed on Multi-PIE dataset, which contains over 750,000 images recorded from 337 subjects using 15 cameras in different directions 963 under various lighting conditions. We take frontal-view images as anchorsand randomly select side-view images (left and right) to form a three view triplet which is the input of our model. In this way, we take 36k training triplets.

\subsubsection{Evaluation Datasets} We mainly perform quantitative and qualitative evaluations on the facescape benchmark containing in-the-wild and in-the-lab data. 14 recent methods are evaluated on the dimensions of camera pose and focal length, which provides a comprehensive evaluation.

\subsubsection{Hyper-parameters setting} In actual training, we set the hyper-parameters in equation~(\ref{eq:newtotal}) to  $\lambda_1 = 1$, $\lambda_2 = 1$,  $\lambda_3 = 1$, $\lambda_4 = 0.5$, $\lambda_5 = 1e-04 $. learning rate $=1e-3$. Train epochs on multi-PIE are 10. \par

\subsection{Qualitative Results}
\label{sec:qualitative}
\begin{figure}[!htp]
\centering
  \includegraphics[scale=0.15]{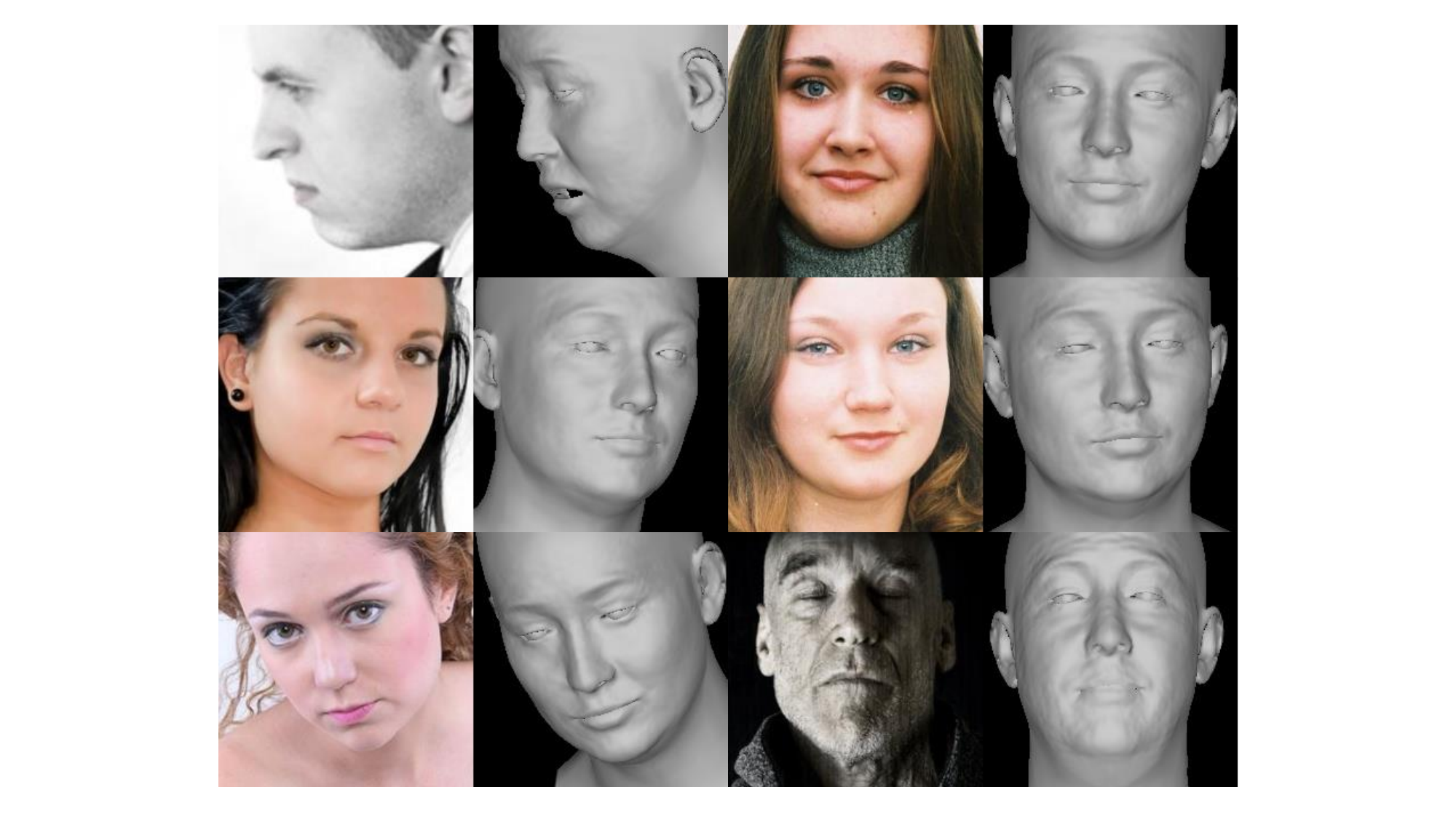}
  \hspace{1cm}
  \includegraphics[scale=0.15]{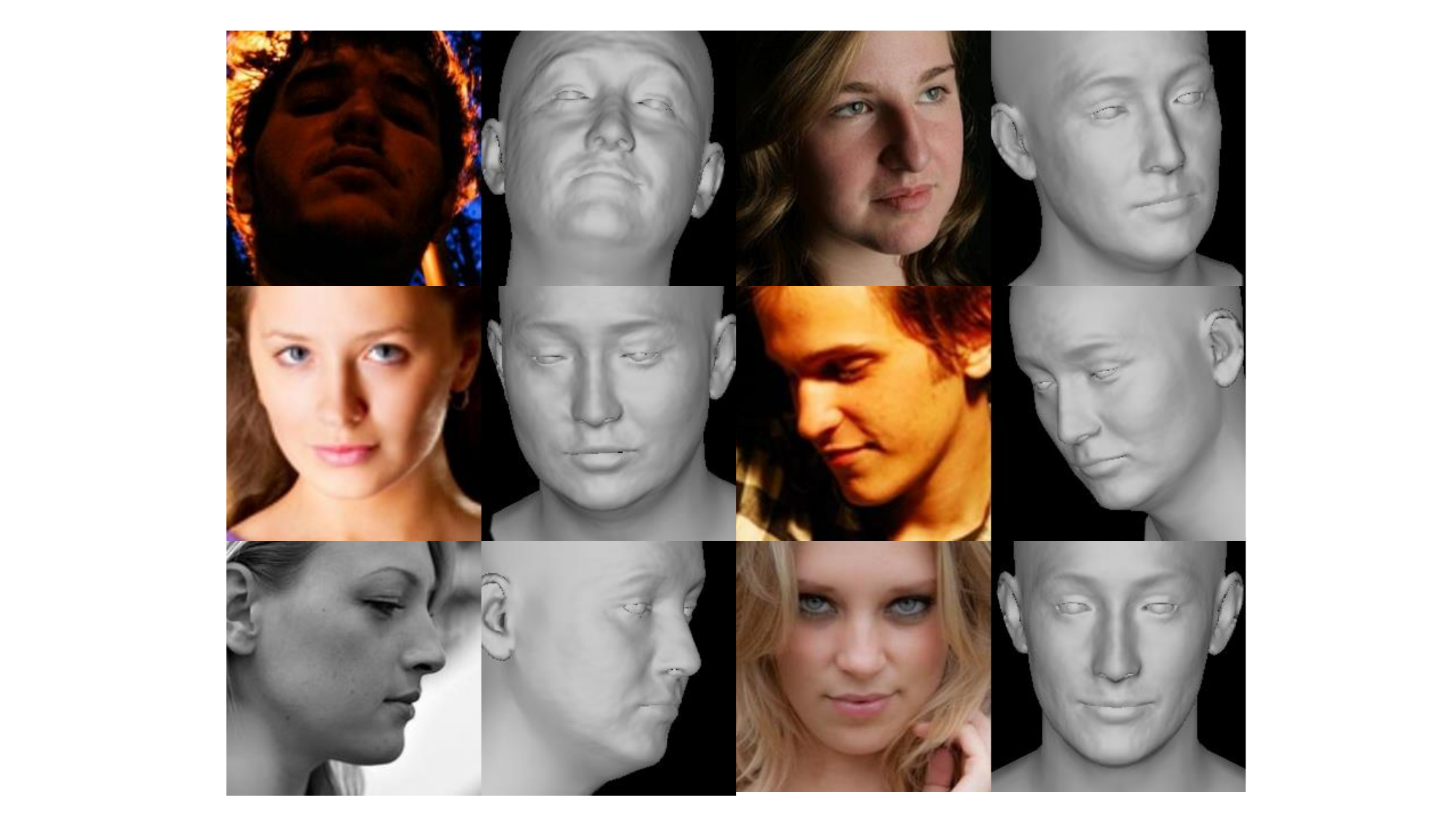}
  % \hspace{0.2cm}
  % \includegraphics[width=0.2\textwidth]{compare.pdf}
  \caption[reconstruction show]
  {MFNet reconstruction. From left to right are input images, MFNet reconstruction.}
  \label{fig:reconshow}
\end{figure}

We first present our reconstruction results, as shown in Figure~\ref{fig:reconshow}. It can be seen that MFNet's reconstructed facial model performs well in various perspectives.

\begin{figure}[!htp]
  % \centering

  \begin{minipage}[t]{0.5\linewidth}
    \centering
    \includegraphics[scale=0.18]{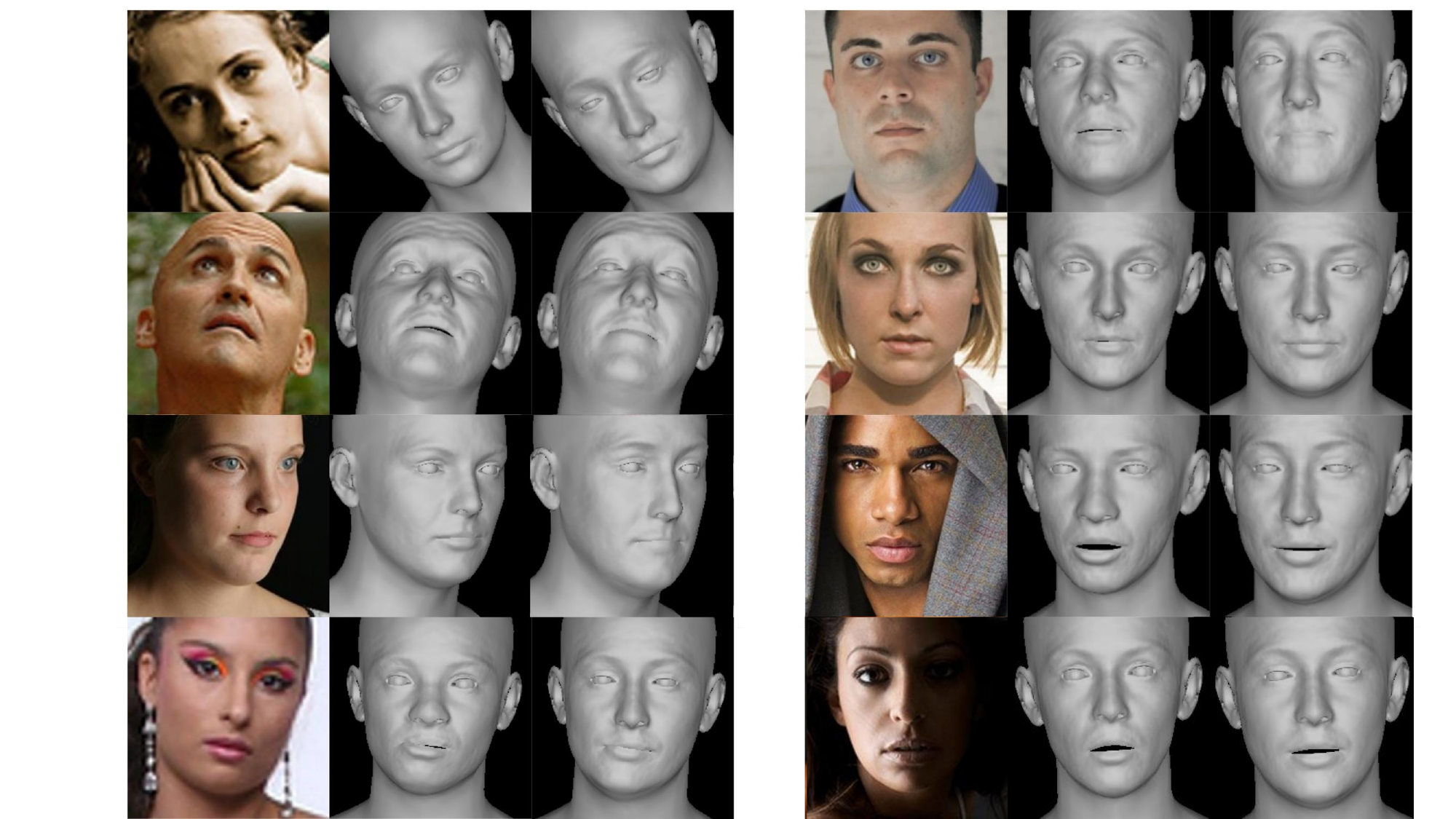}
    \caption{Qualitative experiment of DECA and MFNet.}
    \label{fig:decatoours}
  \end{minipage}
  \begin{minipage}[t]{0.5\linewidth}
    \centering
    \includegraphics[scale=0.18]{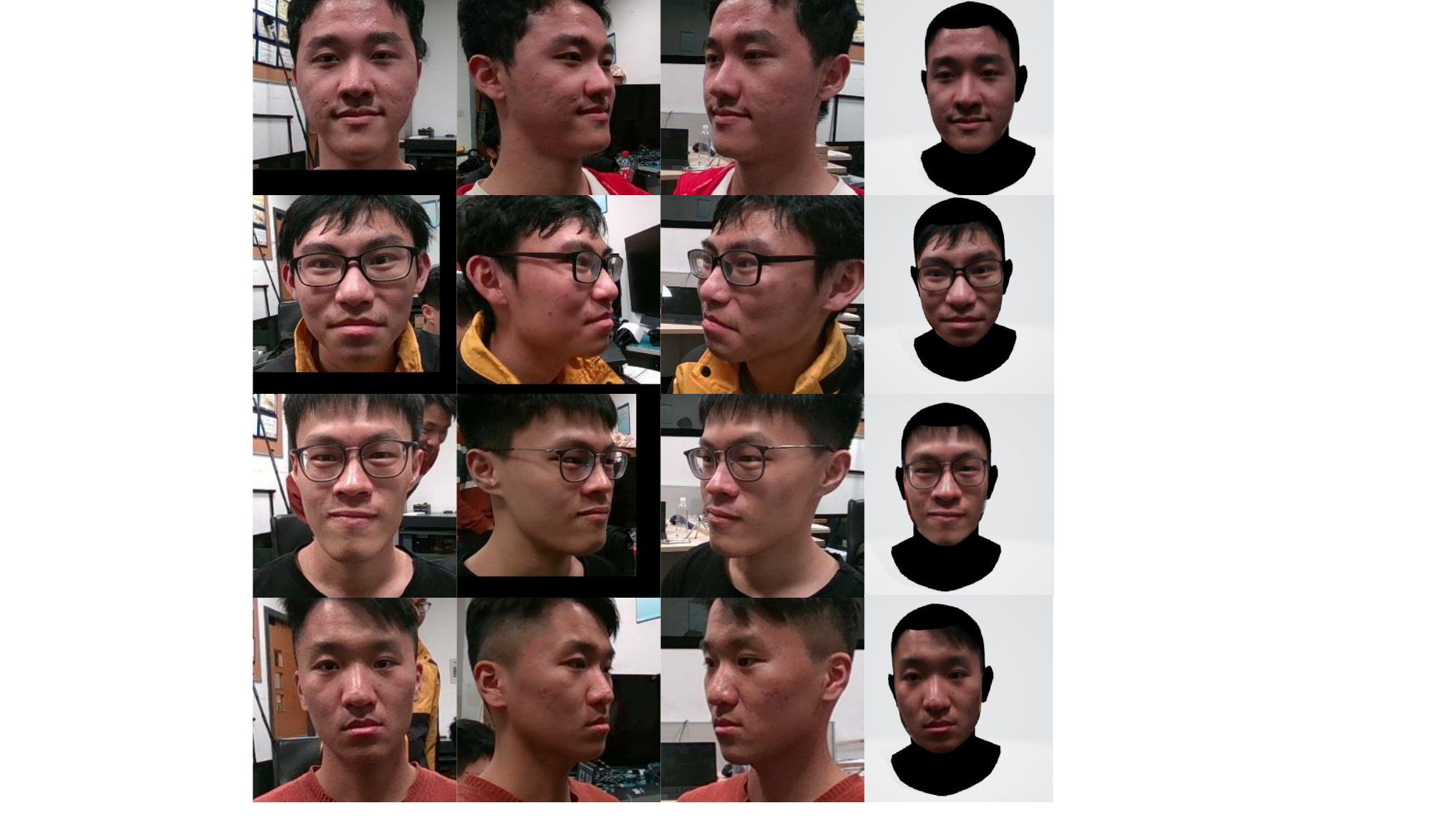}
    \caption{MFNet reconstruction.}
    \label{fig:ourface}
  \end{minipage}

  % \includegraphics[width=0.45\textwidth]{compare.pdf}
  % \hspace{0cm}
  % \includegraphics[width=0.45\textwidth]{ourface.pdf}
  % \caption[MFNet reconstruction of shot images]
  % {Qualitative experiment of DECA and MFNet. From left to right are a certain perspective of the input image, DECA reconstruction, MFNet reconstruction.}
  % \label{fig:ourface}
\end{figure}

Next, we compared the reconstruction results of DECA and MFNet. We used DECA and our model to reconstruct 2000 images from AFLW2000-3D respectively. Some of them are shown in Figure~\ref{fig:decatoours}. Through observation, it can be found that DECA has problems in predicting facial edges in certain situations, but MFNet can reconstruct more accurately due to the involvement of multiple perspectives.

We also set up three-viewed cameras on site to take images of the people around us, obtaining multi-view images that are close to the real environment. We tested the reconstruction effect of MFNet on these images and added texture, as shown in the Figure~\ref{fig:ourface}.

% \begin{figure}[!htp]
%   \centering
  
%   \caption[ourface]
%   {MFNet reconstruction of shot images. From left to right are our faces in three views and the reconstructed face after texture rendering.}
%   \label{fig:ourface}
% \end{figure}

\begin{table*}[t]
  \centering
  \caption[comparison with other single-view methods]
    {Comparison with other single-view methods.}
  \begin{threeparttable}[b]
  \footnotesize
     \begin{tabular}{c | c c c | c c c | c c c | c c c}
      \toprule
      \multirow{2}{*}{methods}  & \multicolumn{3}{c}{0-5} & \multicolumn{3}{c}{5-30} & \multicolumn{3}{c}{30-60} & \multicolumn{3}{c}{60-90}\\
      & CD & MNE & CR & CD & MNE & CR & CD & MNE & CR & CD & MNE & CR \\
      \midrule
      extreme3dface\cite{extreme} & 5.02 & 0.16 & 0.62 & 5.512 & 0.18 & 0.56 & 7.91 & 0.20 & 0.40 & 25.3& 0.26 & 0.27\\
      PRNet\cite{feng2018joint} & 2.61 & 0.12 & 0.83 & 3.11 & 0.11 & 0.83 & 4.25 & 0.11 & 0.78& 3.88 & 0.14 & 0.75\\
      Deep3DFaceRec\cite{deng2019accurate} & 2.30 & \textbf{0.07} & 0.83 & \textbf{2.50} & \textbf{0.07} & 0.83 &3.56 & 0.08& 0.77& 6.81 & 0.14 & 0.62 \\
      RingNet\cite{sanyal2019learning} & 2.40 & 0.08 & \textbf{0.99} & 2.99 & 0.09 & \textbf{0.99} & 4.78 & 0.10& 0.98 & 10.7& 0.18 & 0.97\\
      DFDN\cite{zeng2019df2net} & 3.67 & 0.09 & 0.87 & 3.27 & 0.09 & 0.86 & 7.29 & 0.12 & 0.84 & 27.4 & 0.30 & 0.57 \\
      DF2Net\cite{zeng2019df2net} & 2.92 & 0.12 & 0.57 & 4.21 & 0.13  & 0.56 & 6.54 & 0.15 & 0.46 & 19.7 & 0.30 & 0.30 \\
      UDL\cite{chen2020self} & \textbf{2.27} & 0.09 & 0.69 & 2.59 & 0.09 & 0.68 & 3.45 & 0.10 & 0.64 & 6.32 & 0.17 & 0.49 \\
      facescape\_opti\cite{yang2020facescape} & 2.81 & 0.09 & 0.84 & 3.17 & 0.09 & 0.82 & 4.08  & 0.10 & 0.78  & 6.57 & 0.16 & 0.67\\
      facescape\_deep\cite{yang2020facescape} & 2.70 & 0.08 & 0.87& 3.69 & 0.09 & 0.86 & 4.22 & 0.09 & 0.85 & 9.09 & 0.15 &  0.70\\
      MGCNet\cite{shang2020self} & 2.97 & \textbf{0.07} & 0.84 & 2.94 & \textbf{0.07} & 0.85 & \textbf{2.78} & \textbf{0.07} & 0.81 & 4.20 & \textbf{0.09} & 0.74 \\
      3DDFA\_V2\cite{guo2020towards} & 2.49 & \textbf{0.07} & 0.86 & 2.66 & \textbf{0.07} & 0.86 & 3.17 & \textbf{0.07} & 0.83 & \textbf{3.67} & \textbf{0.09} & 0.79 \\
      SADRNet\cite{ruan2021sadrnet} & 6.60 & 0.11 & 0.90 & 6.87 & 0.11 & 0.89 & 6.39 & 0.10 & 0.84 & 8.62 & 0.16 & 0.82 \\
      LAP\cite{zhang2021learning} & 4.19 & 0.11 & 0.94 & 4.47 & 0.12 &  0.93  & 6.15 & 0.14 & 0.87 & 13.7 & 0.20 & 0.68 \\
      DECA\cite{feng2021learning} & 2.88 & 0.08 & \textbf{0.99} & 2.64 & \textbf{0.07} & \textbf{0.99} & 2.88 & 0.08 & \textbf{0.99} & 4.83 & 0.11 & \textbf{0.99}\\
      MFNet & 3.98 & 0.11 & \textbf{0.99} & 4.07 & 0.11 & \textbf{0.99} & 3.60 & 0.10 & \textbf{0.99} & 5.25 & 0.12 & \textbf{0.99} \\
      % MFNet & & & & & &  \\
      \bottomrule
    \end{tabular}

    \label{tab:multi-measure}
  \end{threeparttable}
\end{table*}
\subsection{Quantitative Results}
\label{sec:quantitative}

At present, there are few benchmarks suitable for multi-view reconstruction test for face parametric models. Therefore, in order to conduct a broader comparison, we test our model on a single view setting and compare it with other algorithms. Due to the original intention of designing MFNet for multi view input methods, this testing method inevitably reduces the reconstruction effect of MFNet. As shown in Table~\ref{tab:multi-measure}, MFNet can not perform best on a single-view testing, but it has already surpassed most models.

To demonstrate the complete performance of MFNet, we also compared it with other models on facescape-lab dataset, which is a multi-view dataset. MFNet used inputs from three views, and others randomly selected one view as input. As can be seen in Table~\ref{tab:Quantitative_exp}, the performance of the complete MFNet model is comprehensively ahead of other models..

\begin{table}[!htpb]
  \centering
    \caption[Comparison of DECA and MFNet]
    {comparison of MFNet and other single-view models. }
    \label{tab:Quantitative_exp}
  \begin{threeparttable}[b]
     \begin{tabular}{c c c c }
      \toprule
      \multirow{2}{*}{method}  & \multicolumn{3}{c}{facescape-lab} \cr
      \cmidrule(lr){2-4} 
	  &CD&MNE&CR\cr
      \midrule
      DECA\cite{feng2021learning} & 5.25 & 0.16 & 0.97 \\
      LAP\cite{zhang2021learning} & 9.76 & 0.20 & 0.85 \\
      SADRNet\cite{ruan2021sadrnet} & 7.21 & 0.18 & 0.89 \\
      DFDN\cite{zeng2019df2net} & 14.10 & 0.32 & 0.93 \\
      Deep3DFaceRec\cite{deng2019accurate} & 5.28 & 0.15 & 0.80 \\
      extreme3dface\cite{extreme} & 15.38 & 0.26 & 0.66 \\
      PRNet\cite{feng2018joint} & 4.97 & 0.15 & 0.85 \\
      facescape\_opti\cite{yang2020facescape} & 5.14 & 0.16 & 0.76 \\
      DF2Net\cite{zeng2019df2net} & 7.39 & 0.17 & 0.67 \\
      MFNet & \textbf{4.89} & \textbf{0.14} & \textbf{0.99}\\
      \bottomrule
    \end{tabular}
  \end{threeparttable}
\end{table}

It can be seen that on the facescape-lab dataset, when MFNet was tested with a complete multi-view input, its various indicators showed significant improvement compared to DECA and also other single-view models, indicating that our multi-view training gives MFNet better reconstruction ability and achieves our expected goals.

\subsection{Ablation Study}
\label{sec:ablation}

In this section, we conduct an ablation study on the
mentioned loss function. In the ablation experiment, we remove one Loss function, keep other Loss function unchanged, and train the same epochs on the same training set. Testing is performed on the fasescape-wild dataset. The results are shown in Table~\ref{tab:loss}. We can see that the whole MFNet has the best performance.

\begin{table*}[!hb]
  \centering
  \caption{Ablation study of loss function.}
  \label{tab:loss}
  \begin{threeparttable}[b]
     \begin{tabular}{c c c c c c c c c c c c c}
      \toprule
      \multirow{2}{*}{methods}  & \multicolumn{3}{c}{0-5} & \multicolumn{3}{c}{5-30} & \multicolumn{3}{c}{30-60} & \multicolumn{3}{c}{60-90} \\
      & CD & MNE & CR & CD & MNE & CR & CD & MNE & CR & CD & MNE & CR \\
      \midrule
      - multiop & 4.29 & 0.12 & 0.98 & 4.43 & 0.12 & \textbf{0.99} & 3.62 & \textbf{0.09} & \textbf{0.99} & \textbf{5.12} & \textbf{0.12} & \textbf{0.99}\\
      - singlelmk & 6.54 & 0.14 & \textbf{0.99} & 5.85 & 0.13 & \textbf{0.99} & 12.2 & 0.18 & 0.97 & 38.6 & 0.25 & 0.93\\
      - eye & 140 & 0.33 & \textbf{0.99} & 423 & 0.38 & 0.98 & 61.8 & 0.24 & 0.96 & 5.91 & 0.14 & \textbf{0.99}\\
      - lip & 6.95 & 0.13 & \textbf{0.99} & 11.2 & 0.15 & 0.98 & 13.7 & 0.17 & 0.94 & 13.6 & 0.18 & 0.95 \\
      - reg & 23.3 & 0.19 & \textbf{0.99} & 32.3 & 0.19 & \textbf{0.99} & 7.39 & 0.12 & \textbf{0.99} & 8.75 & 0.16 & \textbf{0.99}\\
      MFNet & \textbf{3.98} & \textbf{0.11} & 0.98 & \textbf{4.06} & \textbf{0.11} & 0.98 & \textbf{3.60} & 0.10 & \textbf{0.99}  & 5.25 & \textbf{0.12} & \textbf{0.99}\\
      \bottomrule
    \end{tabular}
  \end{threeparttable}
\end{table*}

\begin{figure}[!htp]
  \centering
  \includegraphics[width=0.48\textwidth]{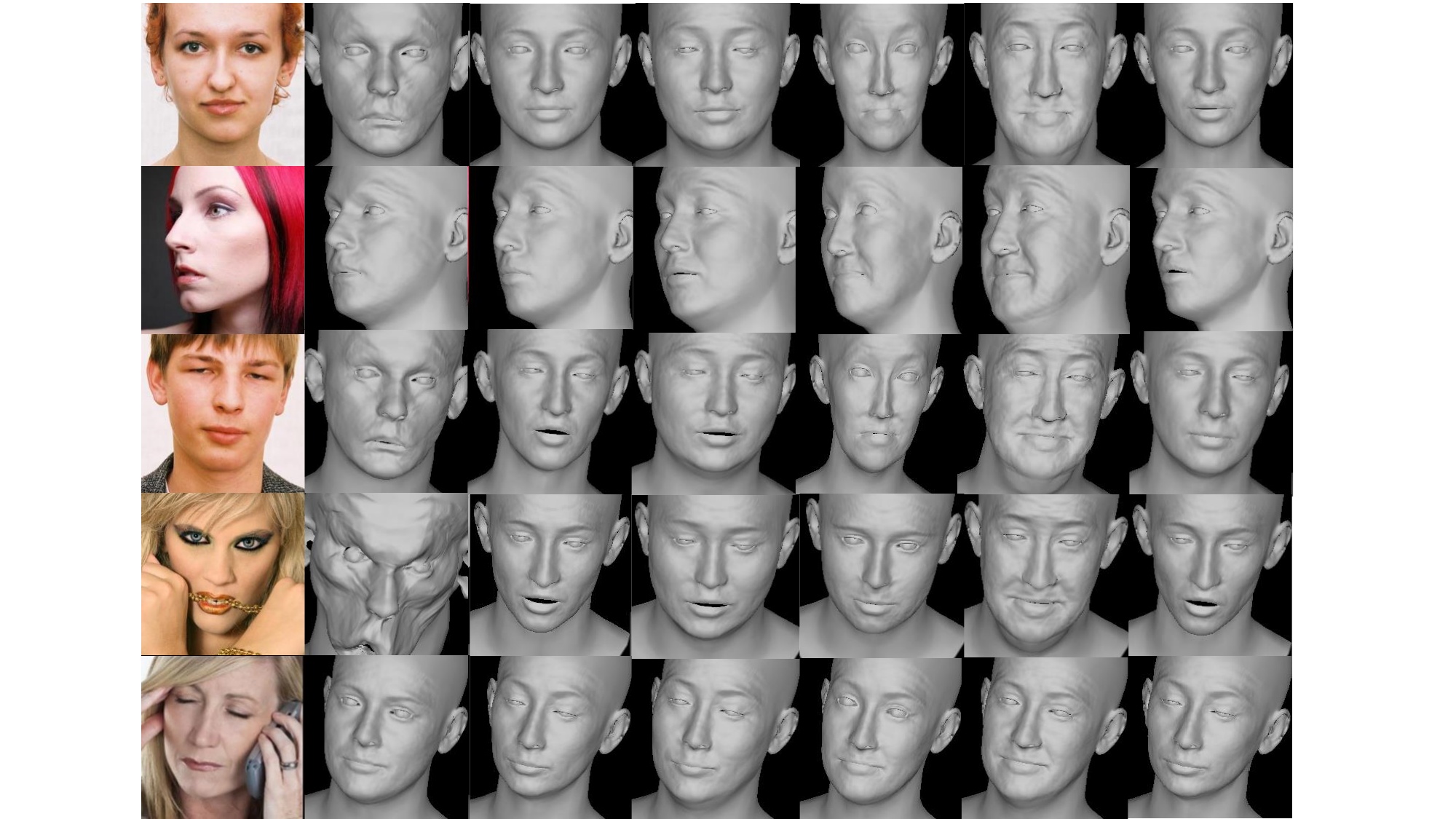}
  \caption[ablation study of loss function]
  {Ablation study of loss function. From left to right are the images with reg, lip, lmk, eye,  multiop removed respectively, and the last column is the reconstruction of MNFet.}
  \label{fig:loss_compare}
\end{figure}

\noindent We also reconstruct each ablation model on the alfw dataset as shown in Figure~\ref{fig:loss_compare}.

\noindent In general, the ablation experiment shows that the performance of the model has declined after the removal of some loss function, which shows that the design of our loss function is reasonable.

\section{Conclusion}

In this paper, we innovatively combine multi-view training with FLAME, propose a multi-view self-supervised framework and implement a complete multi-view training and testing process. Our proposed model MFNet achieve good results on both test datasets and actual captured images. For the implementation of MFNet, we propose a multi-view optical flow loss for our multi-view training framework and propose a novel implementation of the technical details such as covisible mask. Experiments show that our model outperforms other methods in face reconstruction and detail capture, which indicates that the combination of multi-view and FLAME is reasonable.
\bibliographystyle{splncs04}
\bibliography{paper328}

\begin{thebibliography}{10}
\providecommand{\url}[1]{\texttt{#1}}
\providecommand{\urlprefix}{URL }
\providecommand{\doi}[1]{https://doi.org/#1}

\bibitem{blanz1999morphable}
Blanz, V., Vetter, T.: A morphable model for the synthesis of 3d faces. In: Proceedings of the 26th annual conference on Computer graphics and interactive techniques. pp. 187--194 (1999)

\bibitem{blanz2003face}
Blanz, V., Vetter, T.: Face recognition based on fitting a 3d morphable model. IEEE Transactions on pattern analysis and machine intelligence  \textbf{25}(9),  1063--1074 (2003)

\bibitem{booth20173d}
Booth, J., Antonakos, E., Ploumpis, S., Trigeorgis, G., Panagakis, Y., Zafeiriou, S.: 3d face morphable models" in-the-wild". In: Proceedings of the IEEE conference on computer vision and pattern recognition. pp. 48--57 (2017)

\bibitem{booth20163d}
Booth, J., Roussos, A., Zafeiriou, S., Ponniah, A., Dunaway, D.: A 3d morphable model learnt from 10,000 faces. In: Proceedings of the IEEE conference on computer vision and pattern recognition. pp. 5543--5552 (2016)

\bibitem{chen2020self}
Chen, Y., Wu, F., Wang, Z., Song, Y., Ling, Y., Bao, L.: Self-supervised learning of detailed 3d face reconstruction. IEEE Transactions on Image Processing  \textbf{29},  8696--8705 (2020)

\bibitem{deng2019accurate}
Deng, Y., Yang, J., Xu, S., Chen, D., Jia, Y., Tong, X.: Accurate 3d face reconstruction with weakly-supervised learning: From single image to image set. In: Proceedings of the IEEE/CVF conference on computer vision and pattern recognition workshops. pp.~0--0 (2019)

\bibitem{feng2021learning}
Feng, Y., Feng, H., Black, M.J., Bolkart, T.: Learning an animatable detailed 3d face model from in-the-wild images. ACM Transactions on Graphics (ToG)  \textbf{40}(4),  1--13 (2021)

\bibitem{feng2018joint}
Feng, Y., Wu, F., Shao, X., Wang, Y., Zhou, X.: Joint 3d face reconstruction and dense alignment with position map regression network. In: Proceedings of the European conference on computer vision (ECCV). pp. 534--551 (2018)

\bibitem{gerig2018morphable}
Gerig, T., Morel-Forster, A., Blumer, C., Egger, B., Luthi, M., Sch{\"o}nborn, S., Vetter, T.: Morphable face models-an open framework. In: 2018 13th IEEE International Conference on Automatic Face \& Gesture Recognition (FG 2018). pp. 75--82. IEEE (2018)

\bibitem{guo2020towards}
Guo, J., Zhu, X., Yang, Y., Yang, F., Lei, Z., Li, S.Z.: Towards fast, accurate and stable 3d dense face alignment. In: Computer Vision--ECCV 2020: 16th European Conference, Glasgow, UK, August 23--28, 2020, Proceedings, Part XIX. pp. 152--168. Springer (2020)

\bibitem{li2017learning}
Li, T., Bolkart, T., Black, M.J., Li, H., Romero, J.: Learning a model of facial shape and expression from 4d scans. ACM Trans. Graph.  \textbf{36}(6),  194--1 (2017)

\bibitem{loper2015smpl}
Loper, M., Mahmood, N., Romero, J., Pons-Moll, G., Black, M.J.: Smpl: A skinned multi-person linear model. ACM transactions on graphics (TOG)  \textbf{34}(6),  1--16 (2015)

\bibitem{roberts1963machine}
Roberts, L.G.: Machine perception of three-dimensional solids. Ph.D. thesis, Massachusetts Institute of Technology (1963)

\bibitem{ruan2021sadrnet}
Ruan, Z., Zou, C., Wu, L., Wu, G., Wang, L.: Sadrnet: Self-aligned dual face regression networks for robust 3d dense face alignment and reconstruction. IEEE Transactions on Image Processing  \textbf{30},  5793--5806 (2021)

\bibitem{sanyal2019learning}
Sanyal, S., Bolkart, T., Feng, H., Black, M.J.: Learning to regress 3d face shape and expression from an image without 3d supervision. In: Proceedings of the IEEE/CVF Conference on Computer Vision and Pattern Recognition. pp. 7763--7772 (2019)

\bibitem{shang2020self}
Shang, J., Shen, T., Li, S., Zhou, L., Zhen, M., Fang, T., Quan, L.: Self-supervised monocular 3d face reconstruction by occlusion-aware multi-view geometry consistency. In: Computer Vision--ECCV 2020: 16th European Conference, Glasgow, UK, August 23--28, 2020, Proceedings, Part XV. pp. 53--70. Springer (2020)

\bibitem{teed2020raft}
Teed, Z., Deng, J.: Raft: Recurrent all-pairs field transforms for optical flow. In: Computer Vision--ECCV 2020: 16th European Conference, Glasgow, UK, August 23--28, 2020, Proceedings, Part II 16. pp. 402--419. Springer (2020)

\bibitem{tran2018nonlinear}
Tran, L., Liu, X.: Nonlinear 3d face morphable model. In: Proceedings of the IEEE conference on computer vision and pattern recognition. pp. 7346--7355 (2018)

\bibitem{tran2019learning}
Tran, L., Liu, X.: On learning 3d face morphable model from in-the-wild images. IEEE transactions on pattern analysis and machine intelligence  \textbf{43}(1),  157--171 (2019)

\bibitem{extreme}
Trần, A.T., Hassner, T., Masi, I., Paz, E., Nirkin, Y., Medioni, G.: Extreme 3d face reconstruction: Seeing through occlusions. In: Proceedings of the IEEE Conference on Computer Vision and Pattern Recognition. pp. 3935--3944 (2018)

\bibitem{wu2019mvf}
Wu, F., Bao, L., Chen, Y., Ling, Y., Song, Y., Li, S., Ngan, K.N., Liu, W.: Mvf-net: Multi-view 3d face morphable model regression. In: Proceedings of the IEEE/CVF Conference on Computer Vision and Pattern Recognition. pp. 959--968 (2019)

\bibitem{guang2023acc}
Wu, G., Liu, X., Luo, K., Liu, X., Zheng, Q., Liu, S., Jiang, X., Zhai, G., Wang, W.: Accflow: Backward accumulation for long-range optical flow. International Conference on Computer Vision  (2023)

\bibitem{yang2020facescape}
Yang, H., Zhu, H., Wang, Y., Huang, M., Shen, Q., Yang, R., Cao, X.: Facescape: A large-scale high quality 3d face dataset and detailed riggable 3d face prediction. In: IEEE/CVF Conference on Computer Vision and Pattern Recognition (CVPR) (June 2020)

\bibitem{zeng2019df2net}
Zeng, X., Peng, X., Qiao, Y.: Df2net: A dense-fine-finer network for detailed 3d face reconstruction. In: Proceedings of the IEEE/CVF International Conference on Computer Vision. pp. 2315--2324 (2019)

\bibitem{zhang2021learning}
Zhang, Z., Ge, Y., Chen, R., Tai, Y., Yan, Y., Yang, J., Wang, C., Li, J., Huang, F.: Learning to aggregate and personalize 3d face from in-the-wild photo collection. In: Proceedings of the IEEE/CVF Conference on Computer Vision and Pattern Recognition. pp. 14214--14224 (2021)

\end{thebibliography}




% LNCS Homepage, \url{http://www.springer.com/lncs}. Last accessed 4
% \end{thebibliography}
\end{document}